\documentclass[10pt,twocolumn,letterpaper]{article}

\usepackage{iccv}
\usepackage{times}
\usepackage{epsfig}
\usepackage{graphicx}
\usepackage{amsmath}
\usepackage{amssymb}
\usepackage{multirow}
\usepackage{mathtools}
\usepackage{bm}
\usepackage{xcolor}
\usepackage[pagebackref=true,breaklinks=true,letterpaper=true,colorlinks,bookmarks=false]{hyperref}

 \iccvfinalcopy 

\def\iccvPaperID{3468} 

\ificcvfinal\fi

\newcommand{\squeeze}{\vspace{-1em}}
\begin{document}

\title{OSCAR-Net: Object-centric Scene Graph Attention for Image Attribution}


\newcommand\Mark[1]{\textsuperscript{#1}}
\author{Eric Nguyen\Mark{1}\thanks{Equal contribution by co-authors} , Tu Bui\Mark{2}\Mark{*}, Viswanathan Swaminathan\Mark{1}, John Collomosse\Mark{1,2}\\
\Mark{1}Adobe Research, Creative Intelligence Lab — San Jose, CA. \\
\Mark{2}CVSSP, University of Surrey, — Guildford, UK\\
{\tt\small \{enguyen | collomos | vishy\}@adobe.com,  t.v.bui@surrey.ac.uk}
}

\maketitle
\ificcvfinal\thispagestyle{empty}\fi

\begin{abstract}
Images tell powerful stories but cannot always be trusted.  Matching images back to trusted sources (attribution) enables users to make a more informed judgment of the images they encounter online.   We propose a robust image hashing algorithm to perform such matching.  Our hash is sensitive to manipulation of subtle, salient visual details that can substantially change the story told by an image.  Yet the hash is invariant to benign transformations (changes in quality, codecs, sizes, shapes, etc.) experienced by images during online redistribution.  Our key contribution is OSCAR-Net\footnote{Code, data and model available at \href{https://exnx.github.io/oscar/}{https://exnx.github.io/oscar/}} (Object-centric Scene Graph Attention for Image Attribution Network); a robust image hashing model inspired by recent successes of Transformers in the visual domain.  OSCAR-Net constructs a scene graph representation that attends to fine-grained changes of every object’s visual appearance and their spatial relationships.  The network is trained via contrastive learning on a dataset of original and manipulated images yielding a state of the art image hash for content fingerprinting that scales to millions of images.
\end{abstract}



\section{Introduction}

Fake news and misinformation are centuries-old societal problems, exacerbated today by the ease with which digital images are manipulated and shared.  Matching (or `{\em attributing}') an image back to a trusted source, improves user awareness of its origins (or `{\em provenance}') and so enables  more informed trust decisions to be made \cite{ticks,cai}.

Emerging standards for image attribution embed provenance information within metadata \cite{cai,origin}.  Yet image metadata is commonly stripped by social platforms, and may be replaced to misattribute an image \cite{strip}. One solution is to visually match images to a trusted database via `fingerprinting'; \ie visual search or content-aware hashing.

This paper contributes a novel object-centric approach for computing a robust hash from an image, to perform visual matching for image attribution.  Images often undergo {\em `benign transformations'} during online distribution, such as changes in format, resolution, size, padding, \etc. that render cryptographic hashes\cite{sha} unsuitable as fingerprints. Thus image fingerprints must be made robust to these `benign’ transformations. 


\begin{figure}[t!]
    \centering
        \includegraphics[width=1.0\linewidth,height=6cm]{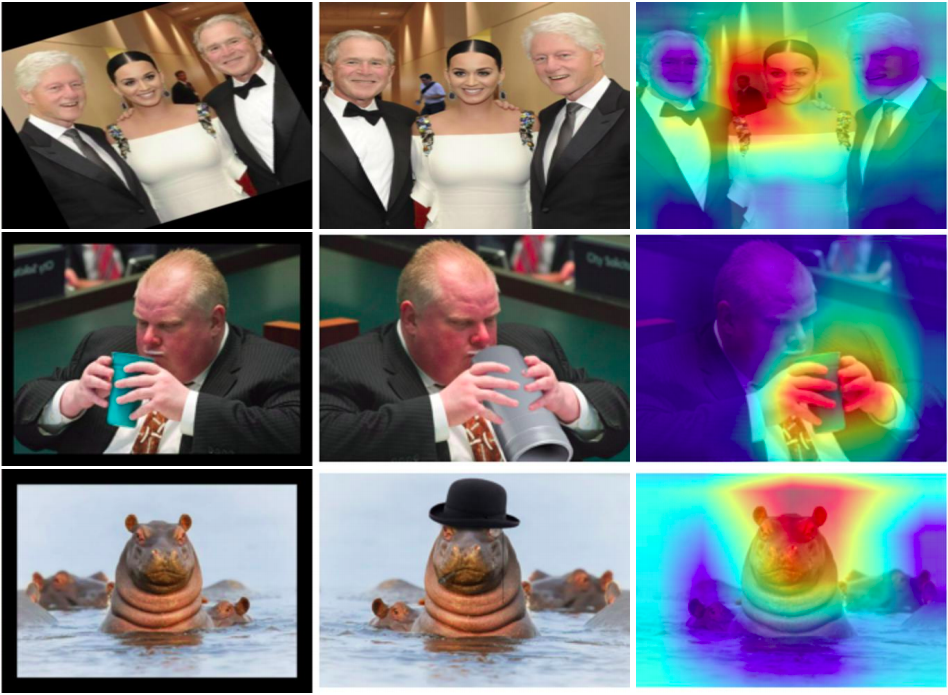}
    \caption{OSCAR-Net learns to create a robust content fingerprint sensitive to subtle manipulations, but invariant to benign transforms. The right columns shows a visualization of regions OSCAR-Net focuses on using Grad-Cam\cite{gradcam}.}
    \vspace{-1em}
    \label{fig:explain}
\end{figure}

We propose a representation learning technique that encourages our hashes to exhibit both invariance to benign transformations and sensitivity to tampering (`manipulation') where image content is altered –- sometimes quite subtly –- but sufficiently to change meaning.  For example, the editorial changes to introduce particular objects/motifs, or alterations to salient visual details such as the face, could substantially change the story told by an image (Fig.~\ref{fig:explain},\ref{fig:psbattles}).  Our method generates substantially different hashes in such cases to avoid corroborating a false story with the provenance information of the original image.  Existing perceptual hashes   \cite{phash,iscc,csq2020cvpr,dsh2016cvpr,dsdh2017nips}) typically exhibit the opposite desired characteristics (c.f. Sec. \ref{sub:scal}), that is, they are often invariant to minor changes, and sensitive to  benign transformations (such as noise, or padding) during content redistribution.

Our core technical contribution is to encode a scene graph representation of the image via a hybrid network architecture, combining a fully-connected Graph Neural Network (GNN) and Transformer to create a robust binary hash of the image.  Our {\em Object-centric Scene Graph Attention for Image Attribution Network} (OSCAR-Net) decomposes a scene into salient objects and aggregates their description into a compact visual hash.  This creates an object-centric hash that exhibits improved sensitivity to minor manipulation of salient objects, and improved robustness to benign transformations.  We leverage contrastive training coupled with benign transformations using data augmentation and manipulations using Adobe Photoshop\textsuperscript{TM} to learn representations that outperform existing perceptual hashing baselines.  Both object appearance and relative object geometry are learned via OSCAR-Net to yield an object-centric hash with improved tamper-sensitivity and scalability.  We combine our  learned OSCAR-Net representation with geometric verification to create a prototype image attribution tool to assist users in determining the provenance of images encountered online.  Such a tool has future potential for fighting fake news by helping users gauge the trustworthiness of content.

\begin{figure*}[t!]
    \centering
        \includegraphics[width=0.9\linewidth,height=6.5cm]{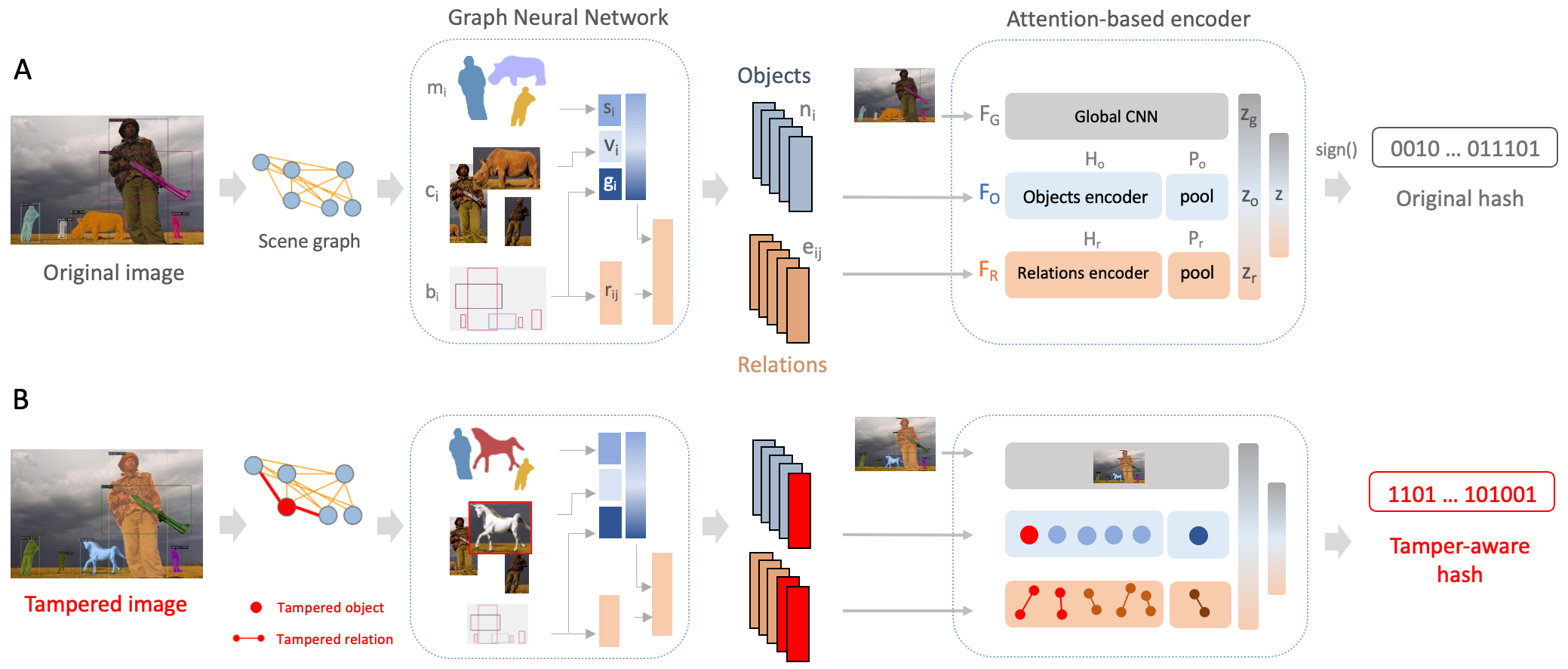} 
    \caption{\textbf{Object-centric Scene Graph Attention Network for Image Attribution (OSCAR-Net)}. In \textbf{A}, our object-centric hashing method first decomposes an image into a fully connected scene graph of $N$ detected objects using their appearance ($c_i$), masks ($m_i$) and bounding box geometry ($b_i$). The whole image, $N$ object and $N^2$ relation embeddings are fed into the 3-stream attention-based encoder, comprised of a global CNN branch $\mathcal{F}_G$, an object encoder $\mathcal{F}_O$ and a relation encoder $\mathcal{F}_R$ to encode each stream to $z_g$, $z_o$, $z_r$, respectively.  A single 64-D embedding $z$ is created, and passed through a $\mathrm{sign}$ function to create a 64 bit image hash.  In \textbf{B}, we indicate how tampered objects and relations are explicitly captured in the scene graph and allow for fine-grained manipulations to produce substantially different hash compared to the original hash.}
    \vspace{-1em}
    \label{fig:pipeline_overview}
\end{figure*}

\section{Related Work}
\label{sec:lit}

\textbf{Visual content authenticity methods} can be categorized into two camps.  The first camp is within digital forensics, and includes techniques such as the `blind' localization of image manipulation \cite{zhang2019}, detection of `deep fake’ (generative) content \cite{kaggledf} based either upon its statistical properties \cite{zhang2020} or its current limitations (\eg absence of blinking \cite{blink}), and anomaly detection \cite{mantranet2019cvpr}. The second camp uses image attribution tools that trace provenance via embedded metadata \cite{cai,origin},  watermarking \cite{hameed2006,devi2009,profrock2006,baba2009}, or  hashing  \cite{iscc,khelifi2017,dsh2016cvpr,hashnet2017iccv,Moreira2018,Zhang2020manip,Bharati2021}. Our work falls in the second camp -- a robust visual hash for image attribution.

\textbf{Perceptual hashing} has been studied extensively for image similarity  \cite{hashsurvey}. Classical approaches sample the spectral domain (\eg DCT \cite{phash,iscc} or wavelet \cite{jacobs1995} co-efficients), or color-texture features \cite{laradji2013,amrutkar2015}. Compact binary codes may be obtained via random projection \cite{lsh} or by learning data-dependent similarity-preserving binary projections such as  product quantization (PQ) \cite{jegou2010product,johnson2019billion}.  More recently, convolutional neural networks (CNNs) have been explored for image similarity. Binary codes may also be learned concurrently with feature learning through two methods: continuous relaxation \cite{dhn2016aaai,dsh2016cvpr,dpsh2016cjai,hashnet2017iccv} or direct optimization in discrete space \cite{dsdh2017nips,dfh2019bmvc,greedy2018nips,dbdh2020}.
As we later show, these methods are vulnerable to subtle tampering on complex images thus not suitable for image attribution applications.

\textbf{Scene graph representation} is an active research area common in scene understanding \cite{moviegraphs2018}, image captioning \cite{guo2019aligning}, retrieval \cite{gcncnn_eccv2020} and synthesis \cite{ashual2019specifying}. Graph neural networks (GNN) are either spectral, which distills information via noise filtering in the graph Fourier domain \cite{kipf2016semi,chen2019multi,scenesketcher}, or non-spectral which operates on the graph directly \cite{guo2019aligning,gcncnn_eccv2020,ashual2019specifying,monet2017}. Non-spectral GNNs can work on graphs with arbitrary structure and various types of nodes and edges, thus can be adapted to complex design elements such as self-attention \cite{gat2018}, autoencoder \cite{gcncnn_eccv2020} and manifold learnings \cite{monet2017}.

To the best of our knowledge, we are the first to leverage scene graph hashing for image attribution. Our contributions are three-fold: (i) we setup new benchmarks for safeguarding visual content integrity via image attribution, (ii) we propose a robust tamper-aware binary fingerprinting approach for images ``in-the-wild" and (iii) we achieve state-of-art performance compared with existing hash methods.

\section{Robust Image Fingerprinting}
\label{sec:method}
Our goal is to learn a model for tamper-aware image hashing that is invariant to benign image transformations, but sensitive to tampering due to subtle yet salient manipulation of content (sec.~\ref{sub:psbattles}).  Since content manipulation often occur at the object level, we explore an object-centric approach that decomposes images into a graph structure which explicitly models their objects and spatial relationships (Fig.~\ref{fig:pipeline_overview}).  Sec.~\ref{sub:graph} describes how we construct a scene graph from an image, sec.~\ref{sub:encoder} describes our scene graph encoder network, sec.~\ref{sub:loss} details our online hashing and losses, and sec.~\ref{sub:train} outlines the training procedure. 

\subsection{Manipulation vs. Benign Transformation}
\label{sub:psbattles}
We trained and evaluated on a dataset called `Photoshopbattles' collected from Reddit forum discussions \cite{psBattles}, which contain manipulated images using Adobe Photoshop\textsuperscript{TM}. This dataset contains {\bf original images} each with several user-manipulated images that often change the story told by the image.  The {\bf manipulated images} include object removal/insertion, face editing/swap, geometry change among other type of manipulations.  We also create {\bf benign transformed images} from each original, applying a suite of primary and secondary transformations such as compression, padding, noise, resampling and affine transformation. More details of this dataset and transformations are discussed in sec.~\ref{sub:train} and \ref{sec:dataset}.


\subsection{Scene Graph Representation}
\label{sub:graph}
We represent an image $x$ by a {\em fully-connected} attributed graph $\mathcal{G}=\{\mathcal{N}, \mathcal{E}\}$, where $\mathcal{N}$ represents node features of the objects in $x$, and $\mathcal{E}$ represents pairwise relationships between every object. We specifically used fully-connected graphs to model any potential tampering between all objects.  To build $\mathcal{G}$, we first identify the objects in an image using MaskRCNN instance segmention~\cite{maskrcnn}. This results in a set of object crops $\{c_i|i=1,2,...,N\}$, where $N$ = number of objects, its corresponding instance segmentation masks $\{m_i\}$, and bounding boxes $\{b_i=[x_i,y_i,h_i, w_i]\}$ characterized by the box center $(x_i,y_i)$ and height/width $(h_i,w_i)$.  We also treat the background (image minus the objects) as its own `object'.  The output of the scene graph representation is $N$ object and $N^2$ relation embeddings.

\textbf{Node features}. We integrate 3 visual cues to construct the node features $\mathcal{N}$.  First, for each object, we extract visual features $v_i$ from the object crop $c_i$.  Specifically, we use $v_i = f^v(c_i)\in \mathbb{R}^{256}$, where $f^v$ is a trainable ResNet50 \cite{he2016deep} with the softmax layer being replaced by a 256-D fully connected layer (training details in sec.~\ref{sub:train}). Second, we compute shape features $s_i \in \mathbb{R}^7$ from the object mask $m_i$ via 7 affine-invariant Hu moments~\cite{humoments}. Third, we use geometry features $g_i\in \mathbb{R}^5$ directly from bounding boxes $b_i$:
\begin{equation}
    g_i = \left[\frac{x_i}{w}, \frac{y_i}{h}, \frac{w_i}{w}, \frac{h_i}{h}, \frac{A_i}{A}\right],
\end{equation}
where $h,w$ are the image height and width; and $A=hw, A_i=x_iy_i$ are the areas of the image and the bounding box respectively. 

We then aggregates $v_i, s_i, g_i$ to create the node features:
\begin{align}
    n_i &= \left[E_v(v_i), E_s(s_i), E_g(g_i)\right]\\
    \mathcal{N} &= \{n_i|i=1,2,...,N \} \in \mathbb{R}^{N\times D_N},
\end{align}
where $E_v, E_s, E_g$ are linear projections of the visual, shape and geometry features; $[,]$ denotes a concatenation; $D_N$ is the resulting feature dimension.

\textbf{Edge features}. We use pairwise geometry relations for edge features in the pixel coordinate space. The geometric connection between the $i^{\textrm{th}}$ and $j^{\textrm{th}}$ objects is defined as:
\begin{equation}
    r_{ij} = \left[\frac{\Delta x}{\sqrt{A_i}}, \frac{\Delta y}{\sqrt{A_i}}, \frac{\sqrt{\Delta x^2 + \Delta y^2}}{\sqrt{w^2+h^2}}, \frac{w_j}{w_i}, \frac{h_j}{h_i}, \theta_{ij}, \gamma_{ij}\right],
\end{equation}
where transition $\Delta x = x_j-x_i$, $\Delta y = y_j-y_i$, box angle $\theta_{ij} = \mathrm{arctan}(\Delta y/\Delta x)$ and standard Intersection over Union (IoU) $\gamma_{ij} = \frac{m_i \cup m_j}{m_i \cap m_j}$. $r_{ij}$ is then used to connect the two objects features:
\begin{align}
    e_{ij} &= \left[n_i, E_r(r_{ij}), n_j\right] \\
    \mathcal{E} &= \{e_{ij} | i,j=1,2,...,N \} \in \mathbb{R}^{N^2\times D_E},
\end{align}
where $E_r$ is another linear layer and $D_E$ is the resulting dimension of the edge features.

\subsection{Scene Graph Encoder}
\label{sub:encoder}
Existing GNN approaches typically define the graph structure based on relationships between objects, creating salient but sparse adjacency matrices. These matrices are acquired primarily either through manual annotations \cite{chen2019multi}, heuristic assumptions about visual similarity \cite{scenesketcher} or geometry \cite{neurocomp19khan,guo2019aligning}, which may not necessarily hold true for ``images in the wild" that have been manipulated.
Instead, we are interested in \emph{all} relationships between objects, since tampering can occur anywhere.  Therefore, we encode graph $\mathcal{G}$ via an attention mechanism that learns object and connection importance between all objects in a fully-connected manner.


\noindent \textbf{The object encoder} $\mathcal{F}_O$ learns an embedding of all objects using node features $\mathcal{N}$ as input. We use a Transformer encoder architecture \cite{vaswani2017attention}, $\mathcal{H}_O$, to implicitly learn the importance of each object's features in an order-invariant manner (we removed the positional encoding so that $\mathcal{N}$ can be treated as an unordered sequence). Here, $\mathcal{H}_O$ has 6 layers each having 8 attention heads. Since the Transformer encoder uses a constant latent vector size for all of its layers, its output sequence has the same length and feature dimension as the input, $H_O(\mathcal{N}) \in \mathbb{R}^{N\times D_N}$. To address this, we add a self-attention module, $P_O$, that acts as a node-wise pooling layer, aggregating the output sequence into a single embedding, following  \cite{sukhbaatar2015end}: 
\begin{align}
    w = \sigma{(\mathrm{tanh}(\mathcal{N}'K^T)V)};\quad
    P_O(\mathcal{N'}) = \sum_i{w_i \mathcal{N}'_i}, \label{eq:selfattn}
\end{align}
where $\mathcal{N'}$ is the Transformer output, the Key $K \in \mathbb{R}^{N\times D_N}$ and Value $V \in \mathbb{R}^N$ behave the same way as the attention mechanism in Transformer but now are learnable parameters. Overall, the Transformer and self-attention layer form our object encoder and together output an object embedding $z_o = \mathcal{F}_O(\mathcal{N}) = P_O(\mathcal{H}_O(\mathcal{N})) \in \mathbb{R}^{D_N}$.

\noindent \textbf{The relation encoder} $\mathcal{F}_R$ uses an identical architecture to $\mathcal{F}_O$ but instead uses edge features as input. $\mathcal{F}_R$ outputs a relation embedding $z_r = \mathcal{F}_R(\mathcal{E}) = P_E(\mathcal{H}_E(\mathcal{E})) \in \mathbb{R}^{D_E}$ where $\mathcal{H}_E$ and $P_E$ are analogous to the object Transformer $\mathcal{H}_O$ and self-attention $P_O$.

\noindent \textbf{Object and global image level fusion} is achieved by fusing our `object-centric' embeddings $(z_o, z_r)$ with a visual appearance feature derived from the entire image, $z_g = \mathcal{F}_G(x)$. $\mathcal{F}_G$ has the same CNN architecture as the object feature extractor $f^v$, but encodes the `global' appearance instead of object crops. We do not share the weights of $\mathcal{F}_G$ and $f^v$ during training. 

Our final embedding is achieved by concatenating the embeddings from the 3 above modules:
\begin{equation}
    z = E_d\left([z_g, z_o, z_r]\right) \in \mathbb{R}^D,
\end{equation}
with $E_d$ being a linear projection layer for a desired output dimension $D$ ($D=64$ in our work). The late-fusion of 3 components - $\mathcal{F}_G$, $\mathcal{F}_O$ and $\mathcal{F}_R$ - allows our model to be sensitive to object level as well as global manipulations.

The closest work to our design is GAT \cite{gat2018} which also
leverages attention mechanism for graph encoding. Our design differs in 3 key aspects: (i) our domain is images
instead of text; (ii) we address a retrieval problem, and require a compact embedding; and (iii) OSCAR-Net uniquely uses 3 encoder streams to encode the global- and object-level features. OSCAR-Net assumes all objects
and relations have connections with each other; the weights of these connections are learned by our Transformer.

\begin{figure}
    \centering
    \includegraphics[width=0.8\linewidth,height=5.8cm]{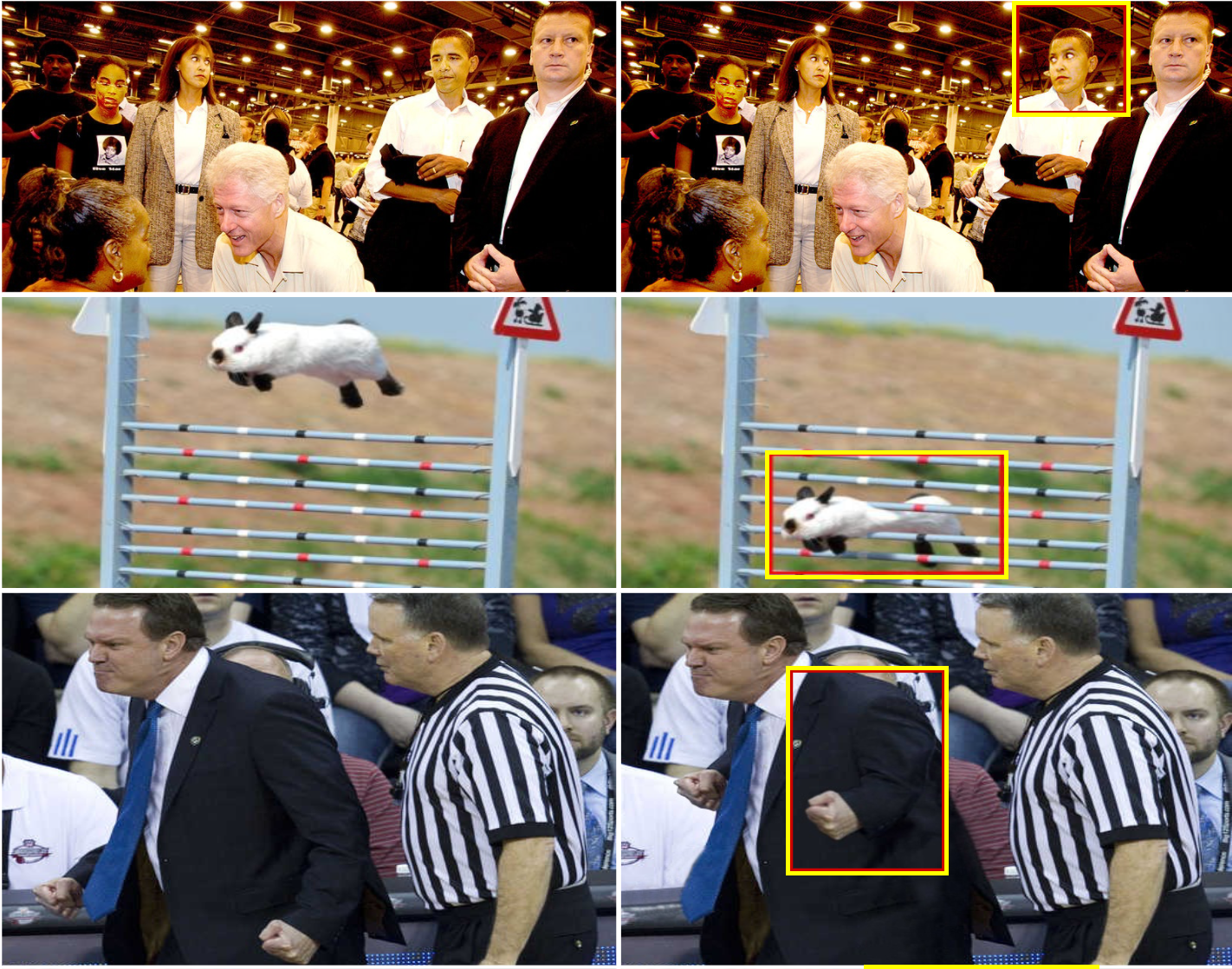}
    \caption{Examples of original (left) and manipulated (right, highlighted) images in PSBattles24K. From the top-down: change in gaze, geometry and pose (see coloured box). More in Supp. Mat.}
    \vspace{-1em}
    \label{fig:psbattles}
\end{figure}

\subsection{Hashing and Losses}
\label{sub:loss}
Hashing is essential for scalable visual search in an image attribution system.  We obtain our hash by discretizing our continuous embedding space z with a $\mathrm{sign}(.)$ layer:
\begin{equation}
    u = \mathrm{sign}(z) \in \{-1, 1\}^D,
    \label{eq:hash}
\end{equation}
The quantization error is approximated via loss:
\begin{equation}
    \mathcal{L}_B(z,u) = \left|\left|z-u \right|\right|^3,
    \label{eq:hash_loss}
\end{equation}
where $\left|\left|.\right|\right|$ denotes entrywise vector norm. Since eq.~\ref{eq:hash} has an ambiguous gradient at zero, we bridge the gradient of $u$ to $z$ during backprop, that is $\partial \mathcal{L}/\partial z = \partial \mathcal{L}/\partial u$, following Discrete Proximal Linearized Minimization (DPLM)  \cite{dplm2016greedy,greedy2018nips} (for more detailed analysis of  eq.~\ref{eq:hash_loss} see Supp. Mat.)

To distinguish manipulation from other benign transformations we propose a supervised version of NTXentLoss/SimCLR \cite{chen2020simple} as the content loss. Under this loss, which we call SimCLR+, each image is pulled towards other benign transformed variants of itself while separating from manipulated variants and other images (different identity) in the batch:
\begin{align}
    \mathcal{L}_C(u_i) &= - \log \frac{\sum_{i+}{\varsigma(u_i,u_{i+})}}{\sum_{i-}{\varsigma(u_i,u_{i-})} + \sum_{j|\psi_j \neq \psi_i}{\varsigma(u_i,u_j)} }
    \label{eq:simclr}
\end{align}
where $\varsigma(u_i,u_j)={\rm e}^{d(E_b(u_i),E_b(u_j))/\tau}$; $E_b$ is a linear layer, $d(.)$ is cosine similarity, and $\tau$ is the temperature for contrastive loss. $E_b$ serves two purposes: (i) mapping the binary code back to a continuous space before the loss computation and (ii) providing an intermediate learning buffer as recommended for SimCLR \cite{chen2020simple}. $\psi_i$ denotes instance class (identity) of image $x_i$, and $u_{i+}$ ({\em resp.} $u_{i-}$) indicates hash code of a benign transformation ({\em resp.} manipulated) image derived from $x_i$. Without $u_{i-}$ the loss degenerates to the standard NTXentLoss and becomes self-supervised.

The total loss becomes $\mathcal{L}(.) = \mathcal{L}_C(.) + \alpha \mathcal{L}_B(.)$ where $\alpha$ is the binary loss weight ($\alpha=10^{-2}$). During inference, $u$ is converted to $\{0,1\}^D$ as $u\coloneqq (u+1)/2$.





\subsection{Training Procedure}
\label{sub:train}
We first train a single CNN model using SimCLR+ loss (eq.~\ref{eq:simclr}), then use it to initialize the weights of both the global CNN ($\mathcal{F}_G$) and object-level CNN ($f^v$) modules. This pretraining step helps to learn a robust encoder for visual appearance, and improves the overall performance (sec.~\ref{sub:abl}). In the main training phase, the whole network is trained end-to-end. We train our models using PyTorch with the ADAM optimizer and step decay learning rate initializing at $10^{-3}$ in  pretraining  and $10^{-4}$ in the subsequent phase. We use an 8GB GTX2080 GPU with a batch size of 20 unique images (prior to data augmentation), and ensure every original image has at least one manipulated version in a batch. 

\textbf{Benign transformations.}
To ensure our model is invariant to benign transformations, we apply a set of transformations on our training images for both the originals and manipulated variants. We define two categories of transforms:  primary and secondary. The primary transforms (including random JPEG compression and resize) are always applied, while 1-3 secondary transforms (horizontal flip, rotation, sharpness, color enhancement, Gaussian noise, padding) are randomly chosen and applied after the primary transforms. This is to reflect that images are commonly compressed and resized when redistributed online.  For details of the parameter ranges for these transformations, see the Supp. Mat. 
\squeeze

\begin{table}[t]
\centering
\begin{tabular}{lcc|cc}
\multirow{2}{*}{Method} & \multicolumn{2}{c|}{Benchmark I} & \multicolumn{2}{c}{Benchmark II} \\
& mAP & mmAP & $F_{mAP}$ & $F_{R1}$ \\ \hline
OSCAR-Net & \textbf{0.8898} & \textbf{0.7411} & \textbf{0.3782} & \textbf{0.3648} \\
GNN & 0.8807 & 0.7111 & 0.3643 & 0.3542 \\
CNN & 0.7980 & 0.6086 & 0.3516 & 0.3308 \\ \hline 

GreedyHash~\cite{greedy2018nips} & 0.6635 & 0.3456 & 0.2367 & \textit{0.2428} \\
HashNet~\cite{hashnet2017iccv} & 0.8093 & \textit{ 0.4031} & 0.2047 & 0.2344 \\
CSQ~\cite{csq2020cvpr} &  0.5785 & 0.2838 & 0.2404 & 0.2356 \\
DFH~\cite{dfh2019bmvc} & 0.3207 & 0.1595 & 0.2118 & 0.1842 \\
DBDH~\cite{dbdh2020} & 0.6908 & 0.3339 & 0.2199 & 0.2268 \\
DSDH~\cite{dsdh2017nips} & 0.6958 & 0.3280 & 0.2078 & 0.2186 \\
ADSH~\cite{adsh2018aaai} & 0.3339 & 0.1887 & 0.1753 & 0.1289 \\
DPSH~\cite{dpsh2016cjai} & 0.8202 & 0.3917 & 0.1724 & 0.2082 \\
DSH~\cite{dsh2016cvpr} & 0.2416 & 0.1358 & 0.1556 & 0.1137 \\
DHN~\cite{dhn2016aaai} & 0.1803 & 0.0898 & 0.1407 & 0.1108 \\
\hline

wHash~\cite{imagehash} & 0.5338 & 0.2274 & 0.0922 & 0.1059 \\
aHash~\cite{imagehash} & 0.5764 & 0.2668 & 0.0919 & 0.1077 \\
pHash~\cite{imagehash} & 0.6008 & \textit{0.3260} & 0.0787 & 0.0972 \\
ISCC~\cite{iscc} & 0.6003 & 0.3252 & 0.0787 & 0.0967 \\
dHash~\cite{imagehash} & 0.6164 & 0.2890 & 0.0604 & 0.0703 \\
cHash~\cite{imagehash}& 0.2509 & 0.1018 & 0.1896 & \textit{0.1674}
\end{tabular}
\caption{Evaluation of proposed method OSCAR-Net versus baselines; all use 64 bit hash codes. }
\label{tab:sota100}
\end{table}

\section{Experiments and Discussion}
\vspace{-1mm}
We evaluate the performance of OSCAR-Net against 6 classical and 10 deep learning baselines for image hashing.

\label{sec:results}
\subsection{Datasets}
\label{sec:dataset}

\begin{table}[t]
\centering
\begin{tabular}{clcc}
& Method & Bm.I mmAP & Bm.II $F_{R1}$ \\ \hline
\multirow{5}{*}{\rotatebox[origin=c]{90}{Content loss}} & SimCLR+ & \textbf{0.7411} & \textbf{0.3648} \\
& Triplet+ & 0.7384 & 0.3627 \\
& SimCLR~\cite{chen2020simple} & 0.7013 & 0.3500 \\
& Triplet~\cite{schroff2015facenet} & 0.7190 & 0.3295 \\
& Pairwise Contrastive~\cite{hadsell} & 0.2199 & 0.1668 \\ \hline 

\multirow{6}{*}{\rotatebox[origin=c]{90}{Hash loss}} & No hashing & 0.8722 & 0.3984 \\
& $\mathrm{sign}()$ (proposed) & \textbf{0.7411} & \textbf{0.3648} \\
& $\mathrm{sign}()$ + Bit Balance & 0.7217 & 0.3582 \\
& $\mathrm{tanh}()$ + Bit Balance~\cite{dbdh2020} & 0.7076 & 0.3556 \\
& $\mathrm{tanh}()$~\cite{hashnet2017iccv} & 0.7096 & 0.3550 \\
& PQ~\cite{jegou2010product} (offline) & 0.6552 & 0.3572
\end{tabular}
\caption{Evaluation of different content (top half) and hashing (bottom half) losses. For content loss experiments, we fix the proposed hash loss $\mathcal{L}_B$ (eq.~\ref{eq:hash}) and vary $\mathcal{L}_C$. Likewise, we fix the proposed content loss $\mathcal{L}_C$ (eq.~\ref{eq:simclr}) and vary $\mathcal{L}_B$ in the hash loss experiments (except for PQ being an offline hashing method).}
\label{tab:loss}
\vspace{-15pt}
\end{table}

\begin{table*}[t]
    \centering
    \setlength{\tabcolsep}{3pt}
    \begin{tabular}{l|cccccccccc}
        Components/Exp. ID & A & B & C & D & E & F (CNN) & G & H (GNN) & I & \textbf{J (full)}\\ \hline 
        Fully connected graph & \checkmark & \checkmark & \checkmark & \checkmark & \checkmark & & \checkmark & \checkmark & & \checkmark \\
        Transformer & \checkmark & \checkmark & \checkmark & \checkmark & \checkmark &  & \checkmark  &  & \checkmark & \checkmark \\
        self-attention & \checkmark & \checkmark & \checkmark & \checkmark & \checkmark &  &  & \checkmark & \checkmark & \checkmark\\
        Visual appearance features $v_i$ & \checkmark & \checkmark & \checkmark & \checkmark & \checkmark & & \checkmark & \checkmark & \checkmark & \checkmark \\
        Vis. appear. feat. w/ pretrain $f^v$ & & \checkmark & \checkmark & \checkmark & \checkmark & & \checkmark & \checkmark &  \checkmark & \checkmark \\
        Geometry feature $g_i$ & & & \checkmark & \checkmark & \checkmark &  & \checkmark & \checkmark & \checkmark & \checkmark \\
        Shape feature $s_i$ & & & & \checkmark & \checkmark &  & \checkmark & \checkmark & \checkmark & \checkmark\\
        Edge feature $e_{ij}$ & & & & & \checkmark &  & \checkmark & \checkmark & \checkmark & \checkmark\\
        Global CNN $\mathcal{F}_G$ & & & & & & \checkmark & \checkmark & \checkmark & \checkmark & \checkmark \\ \hline 
        Benchmark I - mmAP & 0.2715 & 0.6282 & 0.7028 & 0.7015 & 0.7148 & 0.6086 & 0.6597 & 0.7111 & 0.7322 & \textbf{0.7411}\\
        Benchmark II - $F_{R1}$ & 0.1622 & 0.3221 & 0.3547 & 0.3594 & 0.3621 & 0.3308 & 0.3532 & 0.3542 & 0.3599 & \textbf{0.3648}\\
    \end{tabular}
    \caption{Ablation of the proposed (OSCAR-Net) method omitting various architectural stages (see subsec~\ref{sub:abl} for details on Exps. A-J).}
\label{tab:abl}
\vspace{-10pt}
\end{table*}

\noindent {\bf PSBattles24K.} We train and evaluate on a set of user-generated manipulated images posted  released as PSBattles dataset~\cite{psBattles}. The raw PSBattles dataset has 11K original and 90K manipulated variants of those images. To increase the challenge of this dataset, paired images were sorted by ImageNet distance (ResNet50/ImageNet) between the original and manipulated variant, and the lower quartile (most subtle changes) was retained. After duplicate removal, 7K originals and 24K manipulated variants remain (Fig.~\ref{fig:psbattles}). The dataset is split into distinct train and test sets, consisting 21K and 3K pairs of original-manipulated images respectively. 5\% of the training set is for validation.  Further, we create a test set of 150K images (50 random primary and secondary transformations per original) to evaluate invariance to benign transformations (c.f. sec.~\ref{sub:train}).  

\noindent {\bf PSBattles360K-S} We study the robustness of our models toward {\em individual} augmentations using two additional test sets of 10 randomized augmentations applied to each original image within PSBattles24K. The first set contains images from 6 benign transformations seen during the training: compression, rotation, color enhancement, Gaussian noise, padding and sharpness. The second set contain images made from 6 noise sources common in photography and unseen during the training: shot noise, impulse noise, speckle noise, Gaussian blur, de-focus blur and pixelate. Each set has in total 180K ($3K \times 10 \times 6$) images. More details of these transformations can be found in Supp. Mat.

\noindent {\bf Stock4.7M} is a diverse, unannotated dataset of images used to evaluate retrieval performance in the presence of large-scale distractor images. The dataset comprises $4.76M$ images from Adobe Stock website \footnote{\href{https://stock.adobe.com}{https://stock.adobe.com}.  Image list at \href{https://exnx.github.io/oscar}{https://exnx.github.io/oscar}.} at VGA resolution.  Combined with PSBattles24K this enables us to scale evaluation to a test dataset of $\sim 5M$ images.

\subsection{Benchmarks and Evaluation Metrics}
\label{sub:metrics}
To evaluate model performance specifically under the image attribution setting, we defined 2 benchmarks (both are correlated, but differ in terms of the amount of benign vs. manipulated images in the query and database sets).


\textbf{Benchmark I} is used to evaluate embedding separability between manipulated and benign images.  It considers a query set of 3K original images and a search database comprised of (i) 3K manipulated images (ii) 150K benign transformed variants of images from PSBattles24K (iii) 100K distractor images from Stock4.7M.  Given an original image as the query, we wish to retrieve all benign augmentations and reduce the ranking of any manipulated imagery in the returned results.  To this end, we propose a masked Mean Average Precision (mmAP) metric for benchmark I. For a single query q; 
\begin{align}
    mmAP_q &= \frac{\sum_k {r_q(k)m_q(k)P_q(k)}}{\sum_k {r_q(k)}},
\end{align}

where $P_q(k)$ is precision at k, $r_q(k)=1$ if the $k^{th}$ retrieval is relevant otherwise 0, $m_q(k)=1$ if the manipulated image is ranked below k otherwise 0. mmAP thus penalizes early ranking of manipulated images, by computing standard mAP \emph{up to} but excluding the first manipulated result.  mmAP achieves an upper bound of 1 if all benign images are returned on top of the rankings. In scalability experiments where all Stock4.7M distractors are included in the search database, computing full mmAP is not feasible. We therefore use mmAP@$R$ instead; $mmAP_q$ is computed on top $R$ retrieval results only, where $R$ is number of images relevant to query $q$ in the database.   

\textbf{Benchmark II} is aligned closely with our image attribution use case, where the test dataset contains the original images (\eg a database of images with associated provenance information) plus Stock4.7M distractors, and the query set comprises benign and manipulated variants of those originals.  An ideal model would rank an original highly when querying with a benign transformed version, and rank it low for manipulated  queries. Recall at top-1 $R@1$ is computed for benign queries; and $\overline{R@1}=1-R@1$ for manipulated queries. We propose $F_{R1}$ score to measure the trade-off:
\begin{align}
    F_{R1} = R@1 \times \overline{R@1} / (R@1+\overline{R@1}),
\end{align}
We report $F_{mAP}$ in the same manner, computed over mAP.

\subsection{Baseline Comparison}
\label{sub:comparative}
For comparison, we evaluated our OSCAR-Net against the following baselines.  We consider OSCAR-Net and GNN as object-centric, and other baseline methods as ``global" 
\ie encoding the whole image vs decomposing into objects.


\textbf{1. GNN:} We replace the Transformer modules $\mathcal{H}_{O,R}$ of OSCAR-Net with MLP layers to learn graph embeddings while keeping everything else the same. This degenerates to a similar approach to the graph CNNs in \cite{guo2019aligning,gcncnn_eccv2020} although we differ in data domain, losses, and training method.

\textbf{2. CNN:} In contrast with our object-centric approach, this uses a single CNN (ResNet50) to encode the whole image. The model is trained using the same losses and data augmentations detailed in sec.~\ref{sub:loss}-\ref{sub:train}. 

\textbf{3. Classical Methods:} A set of 6 statistical methods for perceptual hashing. Five methods (D-ifference Hash, P-erception Hash, A-verage Hash, W-avelet Hash and C-olor Hash) via public implementations in \cite{imagehash}, also ISCC~\cite{iscc} an ISO standard proposal similar to pHash. All methods produce 64-bit hash codes. 

\textbf{4. Deep Hashing Methods:} A set of 10 deep supervised hashing approaches: CSQ~\cite{csq2020cvpr}, DBDH~\cite{dbdh2020}, DFH~\cite{dfh2019bmvc}, HashNet~\cite{hashnet2017iccv}, GreedyHash~\cite{greedy2018nips}, ADSH~\cite{adsh2018aaai}, DPSH~\cite{dpsh2016cjai}, DSH~\cite{dsh2016cvpr}, DSDH~\cite{dsdh2017nips} and DHN~\cite{dhn2016aaai}. 
We train these models using provided public code using uniform architecture (ResNet50) and code length (64-bit).

\begin{figure*}[t!]
    \centering
        \includegraphics[width=1.0\linewidth,height=4.8cm]{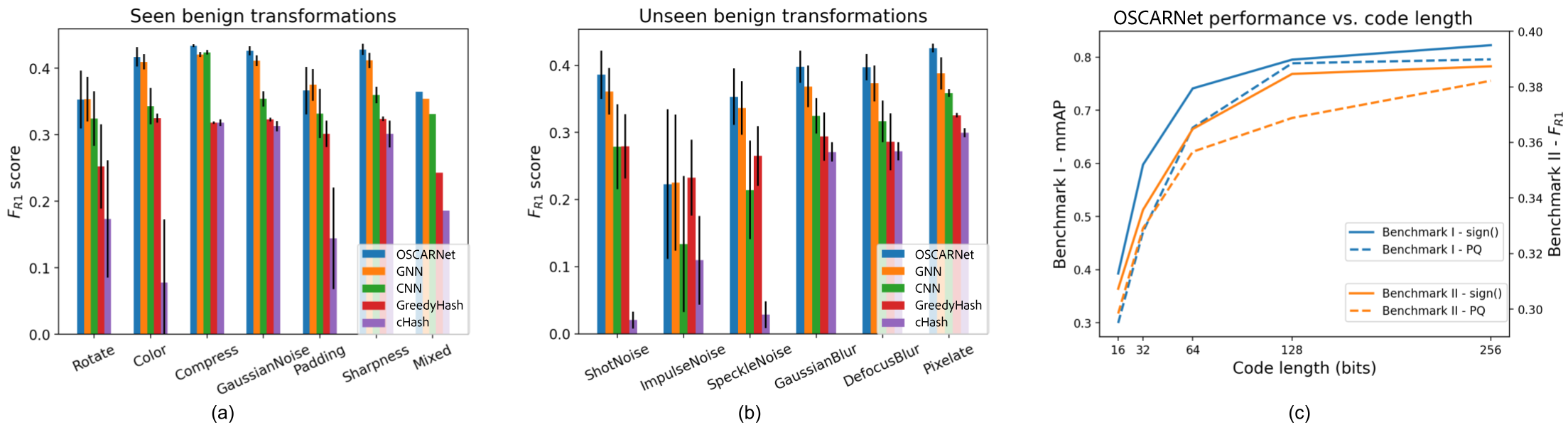} 
    \caption{Characterizing our proposed OSCAR-Net content hashing performance vs. baseline methods.  (a-b): robustness of the image hash to seen/unseen benign transformations;  (c): performance of different hash code lengths for online ($\mathrm{sign}$) and offline (PQ) hashing.}
    \vspace{-1em}
    \label{fig:aug0}
\end{figure*}

Tab.~\ref{tab:sota100} reports the performance of all methods on the PSBattles24K test set with 100K distractor images for benchmark I and II (sec.~\ref{sub:metrics}). Overall the classical methods are at the bottom of the rankings, followed by the global deep hashing methods. The object-centric methods OSCAR-Net and GNN outperform the rest, especially at benchmark I mmAP scores (which takes manipulated images into account as opposed to standard mAP). This indicates the methods' capability in discriminating manipulated content. The object-centric methods also outperform others on benchmark II which requires the balance between benign-transform robustness and manipulation sensitivity. Our proposed OSCAR-Net outperforms all  existing methods and sets a new state-of-the-art in image hash for content fingerprinting.

\subsection{Ablation Studies}
\label{sub:abl}
{\bf Design Component Ablation.} We study the contribution of each design component in OSCAR-Net in Tab.~\ref{tab:abl}. We begin with a minimal design where $\mathcal{N}$ contains only the object features $v_i$ and OSCAR-Net has only 1 encoder stream $\mathcal{F}_O$ (exp.A).  We then add additional components including the pretrained $f^v$ (exp.B), geometry features $g_i$ (exp.C), shape feature $s_i$ (exp.D), the second encoder stream $\mathcal{F}_R$ with edge features $e_{ij}$ (exp.E) to the full model (exp.J). We also remove important components from the full model or test alternative designs (exp.G-I). Pretraining the CNN module $f^v$ is essential as it helps double performance (exp.B vs. A). The geometry features $g_i$ play an important role (exp.C) while the shape $s_i$ and edge $e_{ij}$ features steadily improve performance (exp.D-E). Integrating the global CNN branch $\mathcal{F}_G$ to our object-centric network also boosts performance especially on benchmark I (exp.J), although the single CNN alone does not perform well (exp.F). Exp.G and H show the complementary effects of the transformer $\mathcal{H}_{O,R}$ and self-attention $P_{O,R}$ (eq.~\ref{eq:selfattn}) modules in our network design. We implemented exp.G by removing the self-attention layer and used the feature of the first element in the output sequence as the embedding (similar to \cite{vit2020}), resulting in a performance drop by 8.1\% and 1.2\% on the two benchmarks. Exp.H is the GNN architecture described earlier in sec.~\ref{sub:comparative}, causing 3\% and 1.1\% drop in performance respectively. Additionally, we experimented with a non-fully connected graph in exp.I where two objects are considered as connected only if their relative distance and intersection are within a given threshold, similar to \cite{neurocomp19khan,guo2019aligning}. Non-connected objects are masked out during the attention computation in $\mathcal{H}_O$ and their pairwise relations are removed from the edge features $\mathcal{E}$. Exp.I reports 0.9\% performance drop in benchmark I and 0.5\% on benchmark II. This demonstrates our OSCAR-Net capability in learning the connection weights within a fully connected graph.   

{\bf Content Loss Ablation.} Tab.~\ref{tab:loss} (top) shows how alternative content loss functions affect performance on the two benchmarks. Standard pairwise contrastive loss \cite{hadsell} fails to discriminate subtle differences in manipulated images. Standard SimCLR~\cite{chen2020simple} lags behind by 4\% and 1.5\% on the two benchmarks but still outperforms standard triplet loss~\cite{schroff2015facenet} despite being a self-supervised method. We modified triplet loss by: (i) randomly sampling negative images from other image instances instead of just manipulated images, (ii) randomly feeding the anchor branch with manipulated images and the negative branch with originals, and (iii) adding another L2 term to explicitly push the negatives from the originals. This improved version, named Triplet+, has the closest performance to SimCLR+ but requires a more complex sampling strategy. 

{\bf Hash Loss Ablation.} Tab.~\ref{tab:loss} (bottom) reports on alternative hash functions.  We experimented with $\mathrm{tanh}()$ as a continuous relaxation method \cite{hashnet2017iccv}. We also implemented the bit balancing strategy that regularize the number of 0 and 1 bits, as suggested by \cite{dbdh2020} to improve the hash discrimination. Performance on the 256-D continuous space \ie no hash is reported as an upper-bound reference. Additionally, we compare online hashing with an offline method using Product Quantization (PQ)~\cite{jegou2010product} via FAISS \cite{johnson2019billion}. Offline hash with PQ had the lowest performance (8.6\% and 0.8\% behind our proposed method on the two benchmarks). For online hashing, optimizing directly on the discrete space ($\mathrm{sign}()$) outperforms the relaxation method ($\mathrm{tanh}()$). Also, we find that bit balancing offers only minor improvement for the online approximation method and hinders unnecessary constraints on the proposed direct optimization method.  Overall the use of hashing for image matching enables retrieval speeds of 380ms on average (an indicative speed for 1M images on single-core, i7 CPU) for a 64-bit hash, versus an average of 1.5s for the same using a 256-D real-valued embedding (128x larger footprint). The corresponding difference in accuracy (for our proposed, best case: discrete space optimization) is 13.1\% on benchmark I and 3.4\% on benchmark II justifying the approach.

\subsection{Scalability and Robustness}
\label{sub:scal}

\begin{table}[t]
    \centering
    \begin{tabular}{lcccc}
         Method & 1M & 2M & 3M & 4M \\ \hline
         OSCAR-Net & 0.6680 & 0.6346 & 0.6112 & 0.5930 \\
         GNN & 0.6295 & 0.5949 & 0.5711 & 0.5522 \\
         CNN & 0.5226 & 0.4836 & 0.4619 & 0.4444\\
         HashNet~\cite{hashnet2017iccv} & 0.4064 & 0.3937 & 0.3854 & 0.3785\\
         pHash~\cite{imagehash} &  0.3689 & 0.3689 & 0.3682 & 0.3673
    \end{tabular}
    \caption{Scalability of best performers under Benchmark I:  mmAP@R with increasing  distractors  (Stock4.7M). OSCAR-Net is our proposed method. GNN and CNN are ablations of it.  HashNet and pHash are the benchmark I highest performing deep and classical baseline methods from Tab.1.}
    \vspace{-1mm}
    \label{tab:dis1}
\end{table}

\begin{table}[t]
    \centering
    \begin{tabular}{lcccc}
         Method & 1M & 2M & 3M & 4M \\ \hline
         OSCAR-Net & 0.3247 & 0.3068 & 0.3003 & 0.2952 \\
         GNN & 0.3118 & 0.2944 & 0.2822 & 0.2770 \\
         CNN & 0.3003 & 0.2925 & 0.2880 & 0.2854\\
         GreedyHash~\cite{greedy2018nips} & 0.2284 & 0.2204 & 0.2153 & 0.2105\\
         cHash~\cite{imagehash} & 0.1483 & 0.1464 & 0.1457 & 0.1447
    \end{tabular}
    \caption{Scalability of best performers under Benchmark II: $F_{R1}$ with increasing distractors  (Stock4.7M). GreedyHash and cHash are the benchmark II highest performing deep and classical baseline methods from Tab.1.}
    \vspace{-1em}
    \label{tab:dis2}
\end{table}

\noindent {\bf Scalability}. Tab.~\ref{tab:dis1}-\ref{tab:dis2} show the performance change when up to Stock4.7M images are added to the search database as distractors. We report mmAP@R for benchmark I and $F_{R1}$ for benchmark II, comparing OSCAR-Net, GNN and CNN with the top classic and deep hash performers according to Tab.~\ref{tab:sota100}. Benchmark I shows OSCAR-Net and GNN (ablation) outperforming baselines by a large margin. 
For Benchmark II, GNN underperforms CNN after 2M distractors, potentially due to lack of attention mechanism. This is addressed by OSCAR-Net (via the Transformer module).

\noindent {\bf Transformation robustness}. We study how the types of benign transformations affect the model performance. We set up an experiment in benchmark II style where the benign query set is from PSBattles360K-S. Fig.~\ref{fig:aug0} (a-b) shows $F_{R1}$ performance against seen and unseen transformations. Scores for a subset of transformations used in PSBattles24K are also reported as reference (labeled as ``Mixed"). cHash performs reasonably on blurring/sharpness transformations but is vulnerable to additive noise or color enhancement. The global deep learning methods GreedyHash~\cite{greedy2018nips} and CNN perform better than cHash but consistently underperform the object-centric ones. OSCAR-Net beats other methods at most transformation experiments and  generalizes well for unseen transformations.

\noindent {\bf Code length}. Fig.~\ref{fig:aug0}(c) shows the performance of OSCAR-Net on different code length settings and hashing methods. For offline hashing, we train a OSCAR-Net model outputting 256-D continuous features using the content loss $\mathcal{L}_C$ only, then apply PQ~\cite{jegou2010product} for a desirable code length. For online hashing, we train separate models varying dimension of $z$ and $u$ in eq.~\ref{eq:hash}. Fig.~\ref{fig:aug0}(c) confirms the benefits of online hashing versus offline, since the hash code could be optimized at the same time as its representation is being learned. Also, the optimal code length is shown around 128-bit;  we use 64-bits in our experiments for fair comparison with baselines.

\subsection{Similarity Visualization}

\begin{figure}[t]
    \centering
    \includegraphics[width=0.8\linewidth,height=5cm]{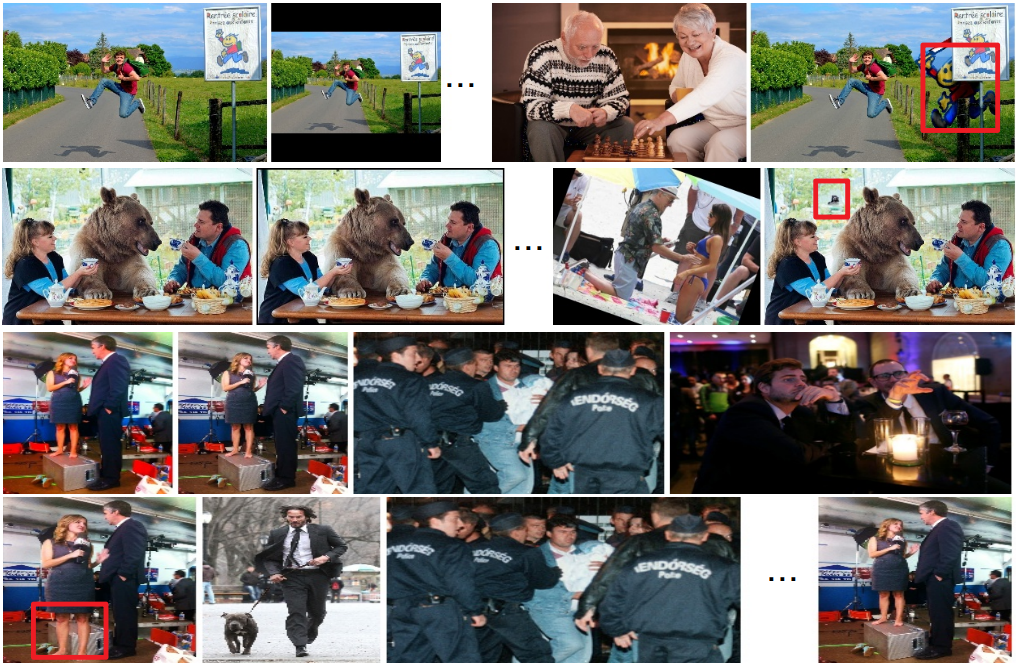}
    \caption{Representative examples (4 rows) of  nearest neighbors in the learned embedding. For this visualization we mix the original, benign and manipulated images.  Originals and benign variants of these are co-proximate, whereas manipulated content (red boxes highlight manipulations) is distant from the original. }
    \vspace{-1em}
    \label{fig:eg}
\end{figure}

We visually explain the image regions that contribute toward  (dis-)similarity, by adapting GradCAM \cite{gradcam} to have objective function:
\begin{equation}
    \mathcal{L}(x | x_+, x_-) = d(f(x),f(x_+)) - d(f(x), f(x_-)),
\end{equation}
where $d(.)$ is cosine similarity. Note that $\mathcal{L}(x | x_+, x_-)$ is similar to a triplet function but with opposite effect. Fig.~\ref{fig:explain} visualizes a heatmap on several original-manipulated image pairs. Regions of interest that contribute most to distance in our embedding are highlighted, regardless of benign transformations (which are learned to be ignored).  Fig.~\ref{fig:eg} includes further  retrieval examples. Manipulations are well separated from originals, but benign images are nearby.

\section{Image Attribution using OSCAR-Net}

\begin{table}[t]
\centering
\begin{tabular}{l|cc|cc|}
\multirow{2}{*}{Method} & \multicolumn{2}{c|}{FPR (\%) $\downarrow$} & \multicolumn{2}{c|}{FNR (\%) $\downarrow$} \\
& 0.1M  & 4M & 0.1M & 4M \\ \hline
OSCAR-Net  & 3.91 & 4.06 & $<1$  & $1.35$   \\
CNN  & 5.41 & 5.71 & $1.01$  & $2.03$  \\
GreedyHash \cite{greedy2018nips}  & 19.36 & 37.43 & 75.86  & 79.51 \\
cHash~\cite{imagehash} & 52.54 & 53.43 & 81.24 &  87.59  \\
\end{tabular}
\caption{Image attribution: False positive/negative rate (FPR/FNR) for proposed (OSCAR-Net) method and baselines (lower is better).  Match via L2 distance in search embedding filtered via geometric verification on a shortlist of the top 10 results. }
\vspace{-1em}
\label{tab:mlesec}
\end{table}

We incorporate OSCAR-Net into a prototype image attribution system where image hashes are used to match images to a database in order to lookup provenance information such as authorship, secure timestamp etc.~\cite{cai}.  Here, images should be matched irrespective of any benign transformations applied to the image during online re-sharing. However, images that have been manipulated should not be matched to their originals in the database to avoid corroborating provenance to a false story.  We conduct retrieval using 10\% of the PSBattles24K query test to identify the top-10 ranked results for a given query, followed by a geometric verification (GV) step.  GV is a common second stage visual search technique used to identify whether any of the top ranked results are relevant or not, based on registration of sparse feature points between the query and result.  We fit a fundamental matrix to homogeneous SIFT point coordinates under MLESAC \cite{mlesac}.  We use GV to avoid the need to specify threshold on distance within the hash embedding; instead relying on pass/fail of registration.

Tab.~\ref{tab:mlesec} reports the false positive rate (FPR) - the fraction of manipulated queries returning originals as a hit; and the false negative rate (FNR) - the fraction of benign queries failing to return originals as a hit. Our OSCAR-Net model exhibits lower FPR/FNR than all deep and classic baseline methods from Tab.~\ref{tab:sota100}.

\section{Conclusion}

We presented OSCAR-Net, an object-centric image hashing method that leverages a paired dataset of original and manipulated images to learn a search embedding for robust visual matching.  Our hash is invariant to  benign transformations typically applied to images distributed online, yet sensitive to fine-grained visual manipulations.  We proposed a hybrid architecture that combined a GNN with Transformers and self-attention.  We showed how our object-centric method significantly improved image search performance and scalability over global (full) image hashing using millions of distractor images.  Future work will integrate our learned embeddings within content provenance systems \cite{cai,origin} to trace images circulating in the wild.  For example, combining our hashes with decentralized immutable storage \eg blockchain and IPFS, in the spirit of recent explorations for digital preservation \cite{archangel,bui2020archangel} but applied to new verticals \eg journalism to combat fake news.

\section*{Acknowledgement}

We thank Andy Parsons, Leonard Rosenthol, Bill Marino and the Adobe Content Authenticity Initiative (CAI) for discussions.  This work was supported by an Adobe Research internship and latterly by DECaDE via EPSRC  Grant Ref EP/T022485/1.

{\small
\bibliographystyle{ieee_fullname}
\bibliography{egbib}
}

\end{document}